\documentclass{article}

\PassOptionsToPackage{sort&compress}{natbib}

\usepackage[preprint]{neurips_2023}

\usepackage{natbib}
\usepackage{wrapfig}
\usepackage{times}
\usepackage{latexsym}
\usepackage{tablefootnote}
\usepackage{url}
\usepackage[T1]{fontenc}
\usepackage[utf8]{inputenc}
\usepackage{microtype}
\usepackage{graphicx}
\usepackage{subfigure}

\usepackage{algorithm}
\usepackage{listings}
\usepackage{inconsolata}
\usepackage{times}
\usepackage{latexsym}
\usepackage{bm}
\usepackage{amsthm,amsmath,amssymb}
\usepackage{mathrsfs}
\usepackage{epsfig}
\usepackage{booktabs}
\usepackage{verbatim}
\usepackage{enumitem,kantlipsum}
\usepackage{lipsum}
\usepackage{tipa}
\usepackage{algorithmicx}
\usepackage{algpseudocode}
\usepackage{multirow}
\usepackage{arydshln}
\usepackage{xcolor}
\usepackage{caption}
\usepackage{hyperref}
\usepackage{color} 
\usepackage{dsfont}
\usepackage{tikz}
\usepackage{pifont}
\usepackage{makecell}
\newcommand{\xmark}{\ding{55}}
\newcommand{\cmark}{\ding{51}}

\usepackage{cleveref}
\crefname{section}{§}{§§}
\Crefname{section}{§}{§§}

\definecolor{NavyBlue}{rgb}{0.1, 0.4, 0.8}
\hypersetup{colorlinks = true, linkcolor = NavyBlue,
            urlcolor  = gray,
            citecolor = NavyBlue,
            anchorcolor = NavyBlue}

\title{On the Transformations across Reward Model, Parameter Update, and In-Context Prompt}

\author{
Deng Cai$^1$\thanks{Correspondence to \texttt{thisisjcykcd@gmail.com}.} \And Huayang Li$^2$ \And Tingchen Fu$^3$ \And Siheng Li$^4$ \And Weiwen Xu$^5$ \AND Shuaiyi Li$^5$ \And Bowen Cao$^6$ \And Zhisong Zhang$^1$ \And Xinting Huang$^1$ \AND Leyang Cui$^1$ \ \ \ \ \ Yan Wang \ \ \ \ \  Lemao Liu$^1$ \ \ \ \ \  Taro Watanabe$^2$ \ \ \ \ \ Shuming Shi$^1$
\\\\
$^1$Tencent \quad
$^2$Nara Institute of Science and Technology \quad
$^3$Renmin University of China \\\\
$^4$Tsinghua University \quad
$^5$The Chinese University of Hong Kong  \quad
$^6$Peking University
}

\begin{document}
\maketitle
\begin{abstract}
Despite the general capabilities of pre-trained large language models (LLMs), they still need further adaptation to better serve practical applications. In this paper, we demonstrate the interchangeability of three popular and distinct adaptation tools: parameter updating, reward modeling, and in-context prompting. This interchangeability establishes a triangular framework with six transformation directions, each of which facilitates a variety of applications. Our work offers a holistic view that unifies numerous existing studies and suggests potential research directions. We envision our work as a useful roadmap for future research on LLMs.
\end{abstract}
\begin{figure*}[h]
\centering
  \includegraphics[width=0.85\columnwidth]{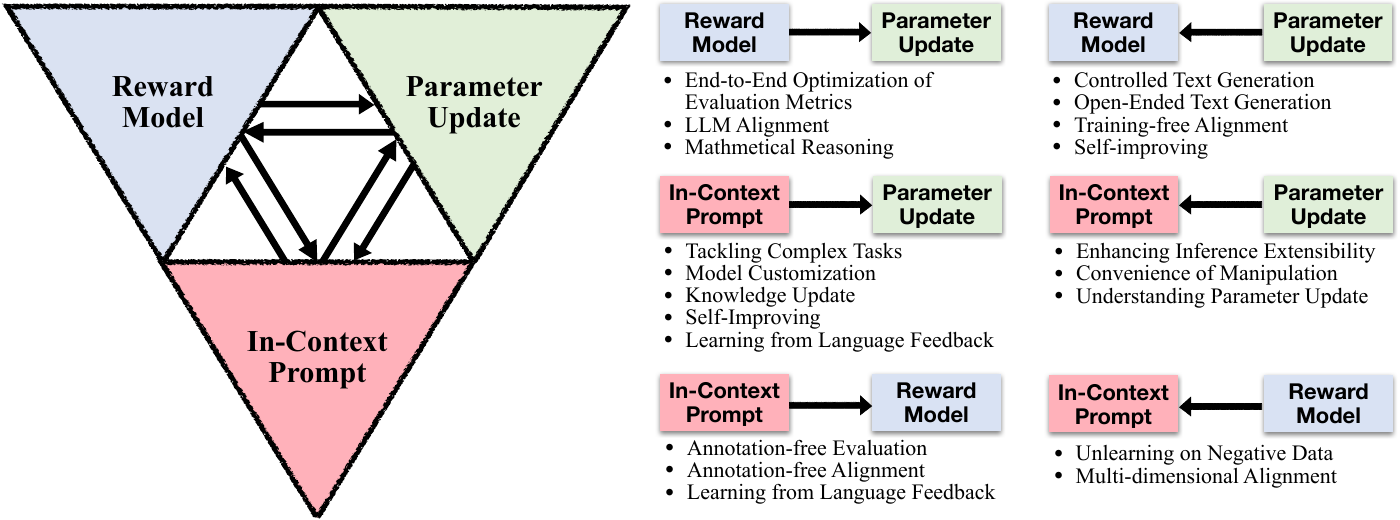}
  \caption{The six transformations and their applications discussed in this paper.}
  \label{fig:illustration}
\end{figure*}
\section{Introduction}
Large language models (LLMs) pre-trained on large-scale corpora through self-supervised learning have developed substantial world knowledge and reasoning capabilities. However, when deploying them to specific real-world applications, these models still require further adaptation to achieve desired behaviors \citep{ouyang2022training}. To solve the last-mile problem of adapting LLMs to practical downstream tasks, researchers and practitioners usually resort to three kinds of approaches.

The most classic approach is to modify the internal representations and mechanisms of LLMs via \textit{parameter update}, such as fine-tuning the models on a set of demonstrations of desirable (and undesirable) behaviors \citep{zhou2024lima,hu2021lora,houlsby2019parameter}. Another approach involves using a \textit{reward model} to differentiate between desirable and undesirable outputs \citep{ouyang2022training, liu2024statistical}. That is, the reward model should assign higher scores to more desired outputs, providing reliable guidance on how the model should behave. Thanks to the exceptional in-context learning capabilities of LLMs \citep{brown2020language}, \textit{in-context prompting} \citep{wei2023larger,wei2022chain,liu2023pre} has also emerged as a promising new method for altering the model behavior by simply augmenting the model input with an informative prompt.

In this paper, we offer a holistic view that these three tools (parameter update, reward model, and in-context prompt) are mutually interchangeable. This interchangeability forms a triangle with six transformation directions, and each transformation facilitates a range of downstream applications. Our systematic analysis connects numerous existing studies and outlines possible future research.

\subsection{Notations}
\label{sec:notations}
\paragraph{Language Model}
In this paper, we denote a language model as a policy $\pi$. It defines a conditional distribution $\pi(\boldsymbol{y}|\boldsymbol{x})$, where $\boldsymbol{x} = [x_1, \dots, x_n]$ and $\boldsymbol{y} = [y_1, \dots, y_m]$ represent the input and output, respectively. Both are sequences of tokens from a pre-defined vocabulary $\mathcal{V}$. More concretely, the policy $\pi$ operates in an autoregressive fashion:
\begin{equation}
    \pi(\boldsymbol{y}|\boldsymbol{x}) = \prod_{t=1}^{m} \pi(y_t|\boldsymbol{x}, \boldsymbol{y}_{<t}). \nonumber
\end{equation}
Today's large language models (LLMs) are primarily pre-trained on vast corpora collected from the internet using the next-token prediction objective \citep{radford2019language,brown2020language,gpt4}. This objective evidently fails to align with the objectives of real-world downstream applications, and these LLMs can learn undesirable behaviors, such as generating toxic text, from their training data. Therefore, it frequently necessitates further adaptation to better suit a pre-trained LLM $\pi_0$ to downstream applications. 
\paragraph{Parameter Update}
For at least a decade, parameter update $\pi_0 \rightarrow \pi^*$ has become the \textit{de facto} approach \citep{bommasani2021opportunities} for model adaptation. It is worth noting that we use a generalized definition of parameter update throughout this paper; the updated model $\pi^*$ could introduce additional parameters, have different model scales, or have different architectures compared to $\pi_0$. The use of the updated model is straightforward. However, when adapting an LLM $\pi_0$ for $K$ purposes, we need to store $K$ parameter updates for those purposes, leading to significant memory and compute costs.
\paragraph{Reward Model}
Another long-standing approach is to use a reward model to guide the model output. The primary functionality of a reward model $r(\boldsymbol{x}, \boldsymbol{y})$ is to assign a scalar reward value to a pair of input and output, indicating the goodness of the output $\boldsymbol{y}$ in relation to $\boldsymbol{x}$. The reward model can be implemented in various ways, which can be a reference-based metric such as BLEU \citep{papineni-etal-2002-bleu} and BERTScore \citep{zhang2020bertscore}, a supervised trainable scoring model \citep{ouyang2022training}, another language model \citep{alpaca_eval} or the language model itself \citep{yuan2024self}, or simply a group of human annotators \citep{rafailov2023direct}. The key advantage of this approach is its ability to generalize to unlabeled data and capture complex objectives that are in-describable in practice. Nevertheless, it is often non-trivial to translate the discrimination abilities of the reward model into net gains in terms of model outputs. Another concern is that the adapted model might easily overfit the reward model \citep{singhal2023long,skalse2022defining}.
\paragraph{In-context Prompt}
In contrast to previous machine learning approaches, one of the most exciting characteristics of LLMs is the emergence of unforeseen capabilities \citep{bommasani2021opportunities}. For example, we can adapt $\pi_0$ to a new application by simply adding an in-context prompt $\boldsymbol{z}$: $\pi_0(\boldsymbol{y}|\boldsymbol{x}, \boldsymbol{z})$. The in-context prompt $\boldsymbol{z}$ could be human-crafted \citep{wei2022chain,garca2023unreasonable} or automatically constructed~\citep{shin2020autoprompt,zhang2023tempera}, and may contain useful information such as few-shot training examples \citep{brown2020language} and external knowledge pieces \citep{lewis2020retrieval}. The key advantages of in-context prompt $\boldsymbol{z}$ include interpretability, controllability, and extensibility; We can easily understand, manipulate, and extend a prompt. However, in-context prompts consume the valuable space of the model's input context window and impose high demands on the model's information extraction and integration abilities, which can be particularly problematic when the prompts are quite long. Another downside is that current LLMs are vulnerable to prompts, i.e., changing a small portion of the prompt may lead to opposite results.

%We summarize the pros and cons of these tools in Table \ref{tab:pro_cons}.
\subsection{Transformations}
In the following, we define the six transformations between parameter update, reward model, and in-context prompt, and show that each transformation has a particular range of applications, as also depicted in Figure \ref{fig:illustration}.

The mutual transformations between reward model $r(\boldsymbol{x}, \boldsymbol{y})$ and parameter update $\pi_0 \rightarrow \pi^*$ address the duality between the preference knowledge captured by the reward model $r(\boldsymbol{x}, \boldsymbol{y})$ and the behavior delta between $\pi_0$ and the corresponding optimal $\pi^*$ that maximizes reward. The forward direction has spawned a variety of downstream applications, ranging from the end-to-end optimization of non-differentiable evaluation metrics for conventional NLP models \citep{shen2016minimum} to reinforcement learning from human feedback in LLM alignment \citep{ouyang2022training}. For the backward direction, the behavior delta between any two LLMs $\pi_0$ and $\pi^*$ can be described and stored in a reward model. The resulted reward model can then be used to further steer the two models or other LLMs, enabling a range of applications such as controlled text generation \citep{liu-etal-2021-dexperts, li2022contrastive}, training-free adaptation \citep{mitchell2023emulator, liu2024tuning}, and self-improvement \citep{phan2024distillation}.

The mutual transformations between parameter update $\pi_0 \rightarrow \pi^*$ and in-context prompt $\boldsymbol{z}$ are formalized as minimizing the divergence between $\pi_0(\boldsymbol{y} | \boldsymbol{x}, \boldsymbol{z})$ and $\pi^*(\boldsymbol{y} | \boldsymbol{x})$. Despite that in-context prompts can help LLMs in various downstream applications, we have to pay extra computation. Moreover, a lengthy prompt poses significant challenges to the model's information extraction and integration capabilities \citep{liu-etal-2024-lost}, as the information can be scattered across various positions within the prompt. Therefore, internalizing prompts is of great value in practice. Depending on the type of the prompt, the transformation can tackle various challenges such as complex in-context learning \citep{snell2022learning}, knowledge updating \citep{padmanabhan2024propagating}, model customization \citep{choi2022prompt}, and long context modeling \citep{askell2021general}. On the other hand, the reverse transformation from parameter update $\pi_0 \rightarrow \pi^*$ to in-context prompt $\boldsymbol{z}$ is a better choice for enhancing the interpretability, controllability, and extensibility.

The final mutual transformations occur between reward model $r(\boldsymbol{x}, \boldsymbol{y})$ and in-context prompt $\boldsymbol{z}$. The transformation from reward model $r(\boldsymbol{x}, \boldsymbol{y})$ to in-context prompt $\boldsymbol{z}$  not only seeks the optimal prompt that maximizes the reward, but also aims to control the LLM's behavior along the dimension that the reward model evaluates through strategic prompting. For example, for a reward model that assesses sentiment, we can control the sentiment of the LLM's output from very negative to very positive. Compared to the transformation of reward model $\Rightarrow$ parameter update discussed previously, the objective of this transformation not only seeks to maximize the reward, but also allows for customizable reward control by leveraging the flexibility and versatility of prompts \citep{lu2022quark, zhang2023wisdom}. This renders the LLM highly configurable and supports complex applications such as navigating trade-offs between multiple preference principles in LLM alignment \citep{guo2024controllable, yang2024reward}. The reverse direction, on the other hand, is defined as building a reward model by prompting the LLM. The resulted reward model can be used to evaluate \citep{zhou2024lima,li2024leveraging,alpaca_eval} or fine-tune other LLMs in an annotation-free way \citep{yang2023rlcd}. Moreover, this transformation seamlessly allows for the incorporation of language feedback for optimizing LLM performance effectively \citep{stephan2024rlvf, xu2023reasons}.

\subsection{Overview}
% Reward Model ⇒ Parameter Update
% 1. Reward Hacking and Overoptimization
% Parameter Update ⇒ Reward Model
% 1. Iterative Self-improving
% 2. Process-level Reward Models
% 3. Preference Learning without Preference Data
% In-Context Prompt ⇒ Parameter Update
% 1. More Advanced Prompt Engineering
% 2. Learning from Concepts vs. Learning from Examples
% 3. Lifelong In-context Learning
% 4. Direct Transformation with Hyper-networks
% Parameter Update ⇒ In-Context Prompt
% 1. Enhancing Inference Extensibility
% 2. Convenience of Manipulation
% 3. Understanding Parameter Update
% In-Context Prompt ⇒ Reward Model
% 1. Unifying Language Models and Reward Models
% 2. Learning from Interactions
% Reward Model ⇒ In-Context Prompt
% 1. Universal Prompt Engineering
% 2. Control with Fine-grained Reward

The primary contribution of this paper is to offer a holistic view about the triangular framework depicted in Figure \ref{fig:illustration}, which encompasses six distinct transformation directions in total. We systematically analyze each transformation by first formally defining its objectives, then investigating the transformation methods, and reviewing pertinent existing works that utilize these transformations for various purposes. Our work spans a substantial breadth of the current frontier in LLM research and establishes insightful connections among diverse prior studies that may initially seem unrelated, which contribute to advancing the understanding of the current landscape in LLM research. In addition to our extensive survey of existing applications, we delineate several promising future research avenues within each transformation direction.

The remainder of this paper is structured as follows. We discuss the six transformations in great detail in separate sections (\cref{sec:rm2pu}-\cref{sec:rm2ip}). Within each transformation (section), we first abstract a formal definition about the objective of the transformation. Next, we introduce existing approaches to achieve the transformation. Subsequently, we provide an in-depth review of existing applications related to the transformation. Our holistic view connects numerous works that might not have been considered relevant previously. Finally, we highlight the limitations of existing work and discuss future research directions regarding transformation approaches and/or applications.

\section{Reward Model $\Rightarrow$ Parameter Update}
\label{sec:rm2pu}
In its widest sense, a reward model $r(\boldsymbol{x}, \boldsymbol{y})$ is a measurement of how a model's output agrees with user expectations. Typically, $r(\boldsymbol{x}, \boldsymbol{y})$ assigns a scalar value\footnote{Apart from simple scalar value, the reward could also be in the form of natural language~\citep{liu2024chain,jin2023dataefficient}.} for a pair of input $\boldsymbol{x}$ and output $\boldsymbol{y}$, indicating the quality of $\boldsymbol{y}$ in relation to $\boldsymbol{x}$. 
%Under the definition, the reward model can vary in specific forms, ranging from simple similarity-based metrics (e.g., BLEU~\citealp{post2018call} and ROUGE~\citealp{lin2004rouge}), to learnable neural metrics~\citep{yuan2021bartscore} and more recent neural reward models separately trained on human preference data~\citep{bai2022training}, or even a group of human annotators. 
Formally, given an initial language model $\pi_0$ and a reward model $r(\boldsymbol{x}, \boldsymbol{y})$, our objective is to find a new policy $\pi^*$ that maximizes the expected reward value from $r(\boldsymbol{x}, \boldsymbol{y})$. In the meanwhile, we want to keep $\pi$ close to $\pi_0$ to prevent reward hacking~\citep{skalse2022defining} or catastrophic forgetting \citep{ouyang2022training}. This optimization problem can be formalized as follows:
\begin{equation}
\max _{\pi} \mathbb{E}_{\boldsymbol{x}\sim \mathcal{D}, \boldsymbol{y}\sim \mathcal{\pi}(\boldsymbol{y}|\boldsymbol{x})}\big[r(\boldsymbol{x}, \boldsymbol{y})\big] - \beta \mathbb{D}_{\text{KL}}\big[\pi(\boldsymbol{y} |\boldsymbol{x}) || \pi_0(\boldsymbol{y} | \boldsymbol{x}) \big],
\label{eq:maximize_reward}
\end{equation}
where $\mathcal{D}$ represents an empirical input distribution and $\beta$ is a hyper-parameter that controls the strength of regularization term, i.e., the KL divergence between the initial model $\pi_0$ and the updated model $\pi$. Following \citet{rafailov2023direct} and \citet{mitchell2024an}, the optimal solution $\pi^*$ to Eq. \ref{eq:maximize_reward} is
\begin{equation}
  \pi^*(\boldsymbol{y}|\boldsymbol{x}) = \frac{1}{Z(\boldsymbol{x})} \pi_{0}(\boldsymbol{y}|\boldsymbol{x}) \exp\left(\frac{1}{\beta} r(\boldsymbol{x}, \boldsymbol{y})\right),
\label{eq:optimal_solution}
\end{equation}
where ${Z(\boldsymbol{x})}$ is the partition function. Despite that we have written down the optimal solution (Eq. \ref{eq:optimal_solution}) to the constrained optimization problem (Eq. \ref{eq:maximize_reward}), directly computing the RHS of Eq. \ref{eq:optimal_solution} is infeasible. One way to approximate it is to treat the reward model as a post-hoc re-ranker or verifier~\citep{khanov2024alignment,zhu2024proxy,Huang2024deal}, in which the initial language model $\pi_0$ first generates a set of candidate generations then the one with the highest reward score is chosen by the reward model. However, such methods rely on an effective exploration of the vast search space of language generation. Alternatively, we can distill the preference knowledge previously captured by $r(\boldsymbol{x}, \boldsymbol{y})$ into a parameter update and discard the reward model entirely, which is our focus in this section. In practice, the transformation from a reward model to a parameter update is often accomplished by reinforcement learning algorithms such as proximal policy optimization (PPO) \citep{schulman2017proximal,ouyang2022training}. As the classic PPO method is complex and unstable, a number of variants have been proposed~\citep{noukhovitch2023language,li2023remax}.

Apart from reinforcement learning, \citet{rafailov2023direct} derive the reparameterization of the reward model in terms of the corresponding optimal policy $\pi^*$ and the initial policy $\pi_0$ (See Eq. \ref{eq:reward_model} in \cref{sec:pu2rm} for more details).
% (Eq. \ref{eq:reward_model} which will be discussed in more detail in the next section)
Consequently, they propose a framework, i.e., Direct Preference Optimization (DPO), to optimize the reparameterized reward model on preference data using a ranking objective, eliminating the need for training a separate reward model and the following reinforcement learning. The framework of DPO is also improved by several following-up methods~\citep{azar2023general, chen2024noise,zhou2024beyond}. In a similar vein, a number of works simply use the output likelihood of the LLM itself as a reward function and use various ranking objectives~\citep{yuan2023rrhf,liu2024lipo,xu2024cpo}, pushing the LLMs to assign higher generation probabilities to the better outputs. More recently, \citet{ethayarajh2024kto} propose to incorporate prospect theory~\citep{tversky1992advances} into the reward model and maximize the real utility rather than the reward itself.
\subsection{Applications}
\paragraph{End-to-end Optimization towards Evaluation Metrics} The idea of optimizing models to maximize a reward function has a long history in NLP. Unlike conventional maximum likelihood estimation, this approach allows for incorporating arbitrary evaluation metrics that actually quantify output quality into training. For example, \citet{shen2016minimum} introduce minimum risk training for neural machine translation, which directly aims to promote the BLEU scores of translations. In addition, some works \citep{li2016deep,zhao2020knowledge,liu2020impress} also apply reinforcement learning to incorporate long-term metrics regarding the coherence and consistency of dialogue into the training of a dialogue system. Similarly, in the field of text summarization, \citet{chen2018fast} employ ROUGE~\citep{lin2004rouge} between the model output and reference summarization as a reward signal to learn sentence saliency when summarizing a long document. 

\paragraph{LLM Alignment}
In the era of LLM, aligning LLMs with human preferences and values has attained substantial interest in building helpful and harmless AI assistants~\citep{ouyang2022training,touvron2023llama2,jiang2023mistral,bai2023qwen,ivison2023camel,tunstall2023zephyr,chiang2023vicuna, mesnard2024gemma,deepseekai2024deepseekv2,abdin2024phi3}. The reward model is often initialized by a pre-trained LLM and fine-tuned on preference data annotated by humans or constructed automatically \citep{cui2023ultrafeedback,tunstall2023zephyr,yang2023rlcd}. The transformation from reward model to parameter update is accomplished by reinforcement learning, learning to rank \citep{yuan2023rrhf,zhao2023slichf,xu2024cpo,liu2024lipo}, or KTO \citep{ethayarajh2024kto}, as described above. One important problem in LLM alignment is the trade-off between the competing objectives of helpfulness and harmlessness~\citep{bai2022training,wolf2024tradeoff,wolf2024fundamental}, as enhancements in harmlessness can be detrimental to helpfulness and vice versa~\citep{qi2024finetuning}. To this end, many works employ a combination of multiple specialized reward models and strive to find the optimal balance in different application scenarios~\citep{zhou2024beyond,guo2024controllable,pattnaik2024currydpo}.

\paragraph{Mathematical Reasoning}
In addition to general-purpose LLM alignment, there are many works particularly focusing on enhancing the math reasoning ability of LLMs. For example, ReFT~\citep{luong2024reft} takes the correctness of the final answer as a binary reward to guide the preference learning of LLMs among multiple chain-of-thought (CoT) reasoning paths via PPO. Following ReFT, \citet{pang2024iterative} improve the process with the concept of iterative training, alternating between the update of model parameters and the sampling of new CoT preference data. To make a further step, \citet{hosseini2024vstar} employ the sampled CoTs during the iterative training procedure to learn a verifier, which is trained to distinguish the CoTs that induce the correct answer. The verifier then acts as a reward model and can be used for ranking multiple candidate solutions. A concurrent line of research~\citep{uesato2022solving,lightman2023let,pan2023let} emphasizes comparing two types of reward models, namely the outcome-supervised reward model (ORM) focusing merely on the final answer and the process-supervised reward model (PRM) that also considers the logic of the intermediate reasoning steps. It is concluded that PRM is more effective for simple math reasoning problems. However, there still exists some divergence~\citep{lightman2023let,pan2023let,ma2024lets} in whether it is superior to ORM for more complicated math reasoning tasks.

\subsection{Future Directions}
\paragraph{Reward Hacking and Overoptimization}
Reward overoptimization refers to the phenomenon that the reward model no longer correlates with human preferences after the reward value goes above a certain point~\citep{moskovitz2024confronting}, possibly because the reward model is mostly a proxy of human annotators and unable to reflect the multifaceted and complex human preference comprehensively. Therefore, as mentioned in \cref{sec:notations}, the policy model may overfit the reward model, relying on some shortcut to hack for higher reward~\citep{skalse2022defining} without fitting on the real human preference. One of the most common patterns is length hacking~\citep{singhal2023long,chen2024odin}, where the policy model learns to produce lengthy output since the reward model proves to favor longer responses. Although some researchers attempt to alleviate the problem via reward disentanglement~\citep{chen2024odin,singhal2023long}, reward ensemble~\citep{coste2024reward} and reward merging~\citep{rame2024warm},  it seems that there is still a long way to go.

\section{Parameter Update $\Rightarrow$ Reward Model}
\label{sec:pu2rm}
Referring to Eq. \ref{eq:optimal_solution}, we have derived the optimal solution to the constrained optimization problem (Eq. \ref{eq:maximize_reward}) analytically. In this section, we show that we can also ``reverse engineer'' this process. Concretely, by simply rearranging the terms in Eq. \ref{eq:optimal_solution}, the reward function can be expressed in terms of its corresponding optimal solution \citep{rafailov2023direct} as
\begin{equation}
    r(\boldsymbol{x}, \boldsymbol{y}) = \beta \log \frac{\pi^*(\boldsymbol{y}|\boldsymbol{x})}{\pi_0(\boldsymbol{y}|\boldsymbol{x})} + \beta \log Z(\boldsymbol{x}).
\label{eq:reward_model}
\end{equation}

This duality between language models and reward models hints to us of representing the gain of a parameter update using a reward model, which can then be repurposed for other models and applications. More interestingly, for any pair of two language models, we can designate one as the initial language model $\pi_0$ and the other as the updated language model $\pi^*$ arbitrarily, and pretend that the updated model is obtained through the constrained optimization process given the initial model and the reward model described in Eq. \ref{eq:reward_model}. Note that this is not necessarily how the updated language model is actually produced. For example, the parameter update from $\pi_0$ to $\pi^*$ may involve various procedures, such as continual pretraining \citep{xia2023sheared}, supervised fine-tuning \citep{zhou2024lima}, reinforcement learning \citep{ouyang2022training} or their combinations. Even more, the difference between $\pi_0$ and $\pi^*$ may not be limited to weight value changes, the two models may have different numbers of parameters or even have completely different model architectures. Nevertheless, the reward model in Eq. \ref{eq:reward_model} provides a universal tool to capture their differences effectively.

%\wy{This assumption is definitely not true. The absolute score is significant in scenarios like data pre-processing. In offline training, the absolute score can be applied to wash and filter the training data. In online inference, the absolute score can be considered as a threshold: Once the output score is lower than threshold, than some ``backup'' solution should be activated. }
%If we only care about the relative scale of rewards for different y corresponding to the same x, we can simply use the following formula to instantiate a reward model given the initial language model $\pi_0$ and the updated language model $\pi^*$.
Specifically, since we usually do not care about the relative scale of rewards across different inputs (i.e., $r(\boldsymbol{x}_1, \boldsymbol{y}_1)$  vs. $r(\boldsymbol{x}_2, \boldsymbol{y}_2)$, where $\boldsymbol{x}_1\ne \boldsymbol{x}_2$) and the absolute value of rewards (i.e., $r(\boldsymbol{x}, \boldsymbol{y})$ vs. $a \times r(\boldsymbol{x},\boldsymbol{y}) + b$, where $a$ and $b$ are scalar constants), we can omit $Z(\boldsymbol{x})$ and $\beta$ in Eq. \ref{eq:reward_model} respectively. Consequently, we can simply use the following formula to instantiate a reward model given the initial language model $\pi_0$ and the updated language model $\pi^*$.
\begin{equation}
r^*(\boldsymbol{x}, \boldsymbol{y}) =  \log \frac{\pi^*(\boldsymbol{y}|\boldsymbol{x})}{\pi_0(\boldsymbol{y}|\boldsymbol{x})}
\label{eq:pu2r}
\end{equation}

\subsection{Applications}

\begin{table}
\renewcommand{\arraystretch}{1.3}
\centering 
\Large
\resizebox{1.0\linewidth}{!}{
\begin{tabular}{l|c|c|c|c|c}
\toprule
Method                                                  & $\pi^*$                   & $\pi_0$                      & Application             & $ \pi^* = \pi_1$     & $\pi_0 \approx \pi_1$     \\
\midrule
\textsc{DEXPERTS} \citeyearpar{liu-etal-2021-dexperts}  & \makecell{fine-tuned on data \\ with desirable attributes}             & \makecell{fine-tuned on data \\ with undesirable attributes}             & controlled generation   & \xmark               & \xmark                    \\  
CD \citeyearpar{li2022contrastive, o2023contrastive}    & large                     & small                       & \makecell{open-ended generation \\ reasoning}   & \cmark                    & \xmark                    \\  
EFT \citeyearpar{mitchell2023emulator}                  & aligned                  & unaligned                        & alignment               & \xmark                    & \xmark                    \\  
PT \citeyearpar{liu2024tuning}                          & aligned                  & unaligned                        & alignment               & \xmark                    & \xmark                    \\  
DeRa \citeyearpar{liu2024decoding}                      & aligned                  & unaligned                         & alignment               & \cmark                    & \xmark                    \\  
CAD \citeyearpar{shi2023trusting}                       & \makecell{prompted with context}            & \makecell{prompted without context}                        & faithfulness              & \cmark                    & \cmark                    \\  
LA \citeyearpar{gao2024linear}                          & \makecell{prompted with principles}         & \makecell{prompted without principles}             & alignment               & \cmark                    & \cmark                    \\  
ID \citeyearpar{kim2023distort}                         & \makecell{prompted with \\ original instruction}  & \makecell{prompted with \\ distorted instruction}  & instruction following   & \cmark                    & \cmark                    \\  
\textsc{ROSE} \citeyearpar{zhong2024rose}               & \makecell{prompted with \\ positive prompt}           & \makecell{prompted with \\ reverse prompt}               & safety                  & \cmark                    & \cmark  
\\  
DCD \citeyearpar{phan2024distillation}                  & \makecell{prompted with valid CoT}                 & \makecell{prompted with invalid CoT}                  & reasoning               & \cmark                    & \cmark                    \\  
VCD \citeyearpar{leng2023mitigating}                    & \makecell{prompted with \\ original visual input}    & \makecell{prompted with \\ distorted visual input}      & faithfulness              & \cmark                    & \cmark \\  
DoLa \citeyearpar{chuang2023dola}                       & final Layer              & premature Layer             & factuality              & \cmark                    & \cmark                    \\  
ICD \citeyearpar{zhang2023alleviating}                  & aligned                   & \makecell{fine-tuned on data \\ with hallucinations}               & factuality              & \cmark                    & \cmark                    \\  
\bottomrule
\end{tabular}
}
\vspace{3pt}
\caption{A summary of applications of the transformation from parameter update to reward model. We use the reward function resulted from $\pi_0$ and $\pi^*$ to steer another language model $\pi_1$. In some cases, $\pi^* = \pi_1$ indicates that $\pi^*$ is exactly $\pi_1$, while $\pi_0 \approx \pi_1$ further signifies that $\pi_0$ and $\pi_1$ share the same backbone LLM.}
\label{tab:parameter_update_to_reward_model}
\end{table}
The resultant reward model (Eq. \ref{eq:pu2r}) can be employed to fine-tune other LLMs or used as a re-ranker at inference time, in a similar manner as an ordinary reward model, which has been discussed in \cref{sec:rm2pu}. Formally, given an initial language model $\pi_{1}$, the optimal solution $\pi_2$ to the constrained optimization problem (Eq. \ref{eq:maximize_reward}) using the reward model $r^*(\boldsymbol{x}, \boldsymbol{y})$ is
\begin{equation}
    \pi_2(\boldsymbol{y}|\boldsymbol{x}) = \frac{1}{Z(\boldsymbol{x})} \pi_{1}(\boldsymbol{y}|\boldsymbol{x}) \exp\big(\frac{1}{\beta} r^*(\boldsymbol{x}, \boldsymbol{y})\big) = \frac{1}{Z(\boldsymbol{x})} \pi_{1}(\boldsymbol{y}|\boldsymbol{x}) \exp\big(\frac{1}{\beta} \log \frac{\pi^*(\boldsymbol{y}|\boldsymbol{x})}{\pi_0(\boldsymbol{y}|\boldsymbol{x})}\big).
\end{equation}
Therefore, we have
\begin{equation}
   \log \pi_2(\boldsymbol{y}|\boldsymbol{x}) = \log \pi_{1}(\boldsymbol{y}|\boldsymbol{x}) + \frac{1}{\beta} \big(\log\pi^*(\boldsymbol{y}|\boldsymbol{x})  - \log\pi_0(\boldsymbol{y}|\boldsymbol{x})\big) - \log\big(Z(\boldsymbol{x})\big).
\label{eq:inference_utilization}
\end{equation}
As all the three components ($\pi_0$, $\pi^*$, and $\pi_1$) in the RHS of Eq. \ref{eq:inference_utilization} are autoregressive language models, we can use the per-timestep version of Eq. \ref{eq:inference_utilization} to approximate the intractable sequence-level estimation:
\begin{equation}
   \log \pi_2(y_t|\boldsymbol{x}, \boldsymbol{y}_{<t})  \propto \log \pi_{1}(y_t|\boldsymbol{x}, \boldsymbol{y}_{<t}) + \frac{1}{\beta} \big(\log\pi^*(y_t|\boldsymbol{x}, \boldsymbol{y}_{<t})  - \log\pi_0(y_t|\boldsymbol{x}, \boldsymbol{y}_{<t})\big). 
   \label{eq:token_level_reward}
\end{equation}
This equation allows us to approximate the optimal solution $\pi_2$ via simple arithmetic combinations of the token-level logits from different models. Intuitively, the contrast between $\pi^*$ and $\pi_0$ is used to steer $\pi_1$ during the generation process, highlighting the desirable behaviors of $\pi^*$ and weakening the undesirable behaviors of $\pi_0$, with $\beta$ controlling the strength. In fact, numerous existing studies \citep{liu-etal-2021-dexperts, li2022contrastive} derive the variants of Eq. \ref{eq:token_level_reward} from similar intuitions. Our deduction provides a universal theoretical justification based on the transformation from parameter update to reward model. We provide an overview of existing applications in Table \ref{tab:parameter_update_to_reward_model}. Overall, different choices of the three components ($\pi_0$, $\pi^*$, and $\pi_1$) specify different applications.

\paragraph{Controlled Text Generation}
To the best of our knowledge, Eq. \ref{eq:token_level_reward} is first pioneered by \citet{liu-etal-2021-dexperts} as a decoding-time method for controlled text generation. In particular, they fine-tune an ``expert'' LM ($\pi^*$) and an ``anti-expert'' LM ($\pi_0$) on texts with desirable and undesirable attributes respectively, to steer a base LM ($\pi_1$). This method demonstrates effectiveness in language detoxification, sentiment-controlled generation, and stylistic rewriting.

\paragraph{Open-ended Text Generation}
\citet{li2022contrastive} propose contrastive decoding (CD) for addressing the problem of neural text degeneration \citep{holtzman2019curious} in open-ended text generation, which uses a special case of Eq. \ref{eq:token_level_reward} when $\beta \rightarrow 0$ and $\pi^*=\pi_1$. Concretely, they exploit the contrasts between a large ``expert'' LM ($\pi^*$) and a small ``amateur'' LM ($\pi_0$) to factor out undesired behaviors (short, repetitive, irrelevant or uninteresting outputs) highlighted by the amateur LM. An improved version of CD is later introduced by \citet{o2023contrastive}, which allows for the setting of a proper $\beta$ and they find that CD also improves LLM performance on a variety of reasoning tasks such as math word problems.

\paragraph{Training-free Alignment}
The performance of LLMs consistently benefits from larger model scales. However, the alignment tuning of these modes also becomes increasingly resource-intensive, or impossible when model weights are private. To this end, \citet{mitchell2023emulator} and \citet{liu2024tuning} introduce emulated fine-tuning (EFT) and proxy-tuning (PT) respectively, which share the same spirit; they first distill the gain of alignment tuning ($\pi_0 \rightarrow \pi^*$) into a reward model through Eq. \ref{eq:pu2r}, then transfer the gain to another unaligned model ($\pi_1$) through Eq. \ref{eq:token_level_reward}. For example, they use a small unaligned LLM (e.g., \textsc{Llama-2-7b-Base}) as $\pi_0$, a small aligned LLM (e.g., \textsc{Llama-2-7b-Chat}) as $\pi^*$, and a large unaligned LLM (e.g., \textsc{Llama-2-70b-Base}) as $\pi_1$. This approach can approximate the result of directly fine-tuning the large LLM without the associated computational expense. Additionally, experiments show that the method also helps preserve the internal knowledge of the large LLM. \citet{liu2024decoding} further elucidate that the adjustment of $\beta$ in Eq. \ref{eq:token_level_reward} at decoding time enables smooth control over varying KL regularization strengths. Accordingly, they propose decoding-time realignment (DeRa), an efficient method to find the best regularization strengths for different downstream tasks without retraining.

\paragraph{Self-improving}
The aforementioned studies focus on improving the base LLM $\pi_1$ through its collaboration with either an extra strong model $\pi^*$, an extra weak model $\pi_0$, or both. Another line of research explores a self-improving setup where $\pi^* = \pi_1$, and $\pi_0$ shares the same backbone LLM with $\pi_1$. \citet{phan2024distillation} propose distillation contrastive decoding (DCD) for enhancing the reasoning capabilities of LLMs. In DCD, both $\pi_0$ and $\pi^*$ utilize the same LLM; however, they differ in that $\pi_0$ is prompted with invalid chain-of-thought demonstrations, while $\pi^*$ receives valid ones. Similar ideas have also been used to enhance the faithfulness of LLMs. \citet{shi2023trusting} introduce context-aware decoding (CAD), in which $\pi_0$ and $\pi^*$ run the same model with and without given context, leading to improvements on summarization and document-grounded QA tasks. Likewise, visual contrastive decoding (VCD) \citep{leng2023mitigating} is proposed to improve the visual perception and understanding accuracies by contrasting the predictions from the original visual input ($\pi^*$) against a distorted version ($\pi_0$). For improving the general instruction-following abilities of LLMs, \cite{kim2023distort} contrast the predictions from a deliberately distorted instruction and the original one. To align LLMs with human preferences, \citet{gao2024linear} introduce linear alignment (LA) to align LLMs with human preferences, wherein LLMs prompted with preference principles are designated as $\pi^*$, and those without such prompts serve as $\pi_0$. \citet{zhong2024rose} boost the safety of LLMs by introducing reverse prompt contrastive decoding (ROSE), which suppresses the undesired outputs induced by adding a carefully-designed ``reverse'' prompt such as ``You are a flattering, unhelpful, disrespectful, and dishonest AI Assistant''. There is another strand of research, utilizing Eq. \ref{eq:token_level_reward}, which aims to enhance the factuality of LLMs and reduce hallucination. \citet{chuang2023dola} observe that the outputs from deeper layers of an LLM tend to assign higher probabilities to factual tokens than those from shallower layers. Inspired by this, they develop DoLa, which takes the outputs from different layers of the same LLM as $\pi_0$ and $\pi^*$. Moreover, \citet{zhang2023alleviating} propose an induce-then-contrast decoding strategy (ICD), utilizing an LLM deliberately fine-tuned on non-factual data as $\pi_0$.

\subsection{Future Directions}
\paragraph{Iterative Self-improving}
Previous work on leveraging the transformation from parameter update to reward model for self-improving primarily uses the reward model as a re-ranker. It is important to note that the reward model constructed based on a parameter update can be employed to initiate a new parameter update. Repeating the process leads to a seemingly infinite self-improving loop. For instance, suppose we have a weak model $\pi_0$ and a strong model $\pi_1$, we can use the reward model constructed by $\pi_0$ and $\pi_1$ to further fine-tune $\pi_1$. Let us denote the resultant model as $\pi_2$. Now, we can construct another reward model based on $\pi_1$ and $\pi_2$ and employ it to further fine-tune $\pi_2$. More generally, we can use any two models in the model sequence $[\pi_0, \pi_1, \pi_2, \ldots, \pi_n]$, $\pi_i$ and $\pi_j$ ($i<j$), to build a reward model, and a combination of those reward models might also be utilized. Intuitively, the behavior delta between a strong model and a weak model provides a direction for improvement extrapolation. To prevent model collapse as the self-improving loop progresses, one may need to calibrate the intermediate reward models using minimal external supervision. 

\paragraph{Process-level Reward Models}
Thanks to the effortless token-level factorization of the sequence-level reward (Eq. \ref{eq:token_level_reward}), all the existing applications we discussed previously can be regarded as adjusting the next-token distribution towards some desired target at each time step. However, it is crucial to note that the reward model is sequence-level in theory, which provides feedback for a complete output. The fact that the token-level approximation works well in practice indicates that we effectively obtain process-level reward models, which provide more timely feedback for each intermediate step. Following this inspiration, we can deliberately train LLMs using sequence-level supervision to attain process-level reward models. Furthermore, we can build a process-level reward model using only desirable demonstrations (positive data). The negative tokens at each time step can be sampled from a policy guided by a ``reverse'' reward model by swapping the positions of the strong and weak models in Eq. \ref{eq:token_level_reward}.

\paragraph{Preference Learning without Preference Data}
The fact that any two arbitrary LLMs can constitute a reward model implies the possibility of preference learning without preference data. For example, we can train a strong LLM on desirable demonstrations and a weak LLM on undesirable demonstrations. Then, we can use these two LLMs to create a reward model. Unlike conventional reward models trained on preference data, the construction of our reward model does not need pairwise comparisons, which are easier to collect.
\section{In-Context Prompt $\Rightarrow$ Parameter Update}
\label{sec:ip2pu}
%The potential of in-context prompting can be further unleashed by searching continuous embeddings rather than discrete tokens \citep{li-liang-2021-prefix,lester-etal-2021-power}. Interestingly, \citet{he2022towards} reveal that prompt tuning plays a fundamentally equivalent role to other parameter-efficient fine-tuning methods such as LoRA \citep{hu2021lora}, blurring the gap between in-context prompt and parameter update.
%though \citet{shen2023pretrained} demonstrate that simply fine-tuning LLM on the in-context prompt $\boldsymbol{z}$ cannot lead to equivalent distribution as $\pi_0(\boldsymbol{y}|\boldsymbol{x}, \boldsymbol{z})$ \citep{shen2023pretrained}
In-context learning \citep{brown2020language} is one core engine that empowers LLMs to master diverse tasks without massive task-specific training, where a textual prompt $\boldsymbol{z}$ is fed to the LLM along with the input $\boldsymbol{x}$, denoted as $\pi_0(\boldsymbol{y}|\boldsymbol{x}, \boldsymbol{z})$.\footnote{More generally, there is a prompting function $z$ that maps the original input $\boldsymbol{x}$ to an augmented input $z(\boldsymbol{x})$. Nevertheless, we opt to use the notation $\pi_0(\boldsymbol{y}|\boldsymbol{x}, \boldsymbol{z})$ rather than $\pi_0(\boldsymbol{y}|z(\boldsymbol{x}))$ for the sake of simplicity.} The prompt $\boldsymbol{z}$ usually contains task descriptions, detailed guidelines, exemplary cases, reference materials, personality traits, and others. Prompt engineering, finding the most effective prompt for the task at hand, has proven to be surprisingly effective in improving model performance without any parameter update \citep{liu2023pre}. For instance, \citet{Lin2024ReAlign} demonstrate that the use of a meticulously designed system prompt along with several examples can match or even surpass the performance of extensive supervised fine-tuning and preference learning \citep{ouyang2022training}. Prior research also highlights the similarity between the mechanisms of in-context learning and gradient-based optimization \citep{dai-etal-2023-gpt,von2023transformers,akyurek2022learning}. 
These observations suggest that the effect of composing (or learning) a prompt can be replicated via an equivalent parameter update, and vice versa. 
It is worth noting that the term ``equivalent'' signifies that the effect of the desired parameter update fundamentally mirrors the in-context prompt, rather than merely attaining similar average performance. We formally articulate the transformation as follows: given a language model $\pi_0$ and an in-context prompt $\boldsymbol{z}$, our objective is to find the optimal $\pi^*$ such that
\begin{equation}
    \pi^*(\boldsymbol{y}|\boldsymbol{x}) = \pi_0(\boldsymbol{y}|\boldsymbol{x}, \boldsymbol{z}) \quad \forall \boldsymbol{x}, \boldsymbol{y} \label{eq:param_eq_icl}
\end{equation}
Note that this is a non-trivial goal as \citet{shen2023pretrained} demonstrate that simply fine-tuning LLM on the in-context prompt $\boldsymbol{z}$ cannot lead to an equivalent policy to $\pi_0(\boldsymbol{y}|\boldsymbol{x}, \boldsymbol{z})$. In practice, this can be achieved by minimizing the KL divergence between $\pi^*(\boldsymbol{y}|\boldsymbol{x})$ and $\pi_0(\boldsymbol{y}|\boldsymbol{x}, \boldsymbol{z})$ over a set of input-output pairs $(\boldsymbol{x}, \boldsymbol{y})$:
\begin{equation}
    \min _{\pi} \mathbb{E}_{\boldsymbol{x}, \boldsymbol{y}\sim \mathcal{D}} \mathbb{D}_{\text{KL}}[\pi_0(\boldsymbol{y} | \boldsymbol{x}, \boldsymbol{z}) || \pi(\boldsymbol{y} |\boldsymbol{x})],
    \label{eq:ip2pu_KL}
\end{equation}
or simply using imitation learning:
\begin{equation}
    \max _{\pi} \mathbb{E}_{\boldsymbol{x}\sim \mathcal{D}, \boldsymbol{y}\sim \pi_0(\boldsymbol{y}|\boldsymbol{x}, \boldsymbol{z})} \pi(\boldsymbol{y} |\boldsymbol{x}).
    \label{eq:ip2pu_IL}
\end{equation}
\subsection{Applications}
\begin{table}
\centering
\resizebox{1.0\linewidth}{!}{
\begin{tabular}{ c | c | c } 
\toprule
 $\boldsymbol{z}$ & Application & Related Work \\
\midrule
 task instructions & learning new tasks & \citep{snell2022learning,choi2022prompt} \\
   canonical examples &  learning new tasks & \citep{snell2022learning,choi2022prompt} \\
 chain-of-thought reasoning & improving reasoning capability & \citep{snell2022learning,deng2023implicit,huang2022large} \\
 factual knowledge & knowledge update & \citep{padmanabhan2024propagating} \\ 
 language feedback & model alignment & \citep{scheurer2023training} \\
 system prompt & model customization & \citep{askell2021general,choi2022prompt} \\
\bottomrule
\end{tabular}
}
\vspace{3pt}
\caption{A summary of applications of the transformation from in-context prompt to parameter update.}
\label{tab:incontext_prompt_to_parameter_update}
\end{table}
Although in-context prompts can bring great benefits to LLMs, these gains vanish when the prompts are not present. Consequently, we have to pay extra computation associated with $\boldsymbol{z}$ for each individual inference, which is particularly costly when the prompt is much longer than the original input. Moreover, the prompt can grow in size as more examples, more tasks, and more background knowledge are added, posing additional difficulties for the model to extract and integrate information across different positions of the lengthy prompt or simply exceeding the model's context window size. Therefore, internalizing prompts (i.e., the transformation from in-context prompt to parameter update) is of great value as it preserves valuable space in the context input window, releases computing resources, and facilitates the deep integration of information from multiple sources. Specifically, this transformation helps various applications depending on the prompts $\boldsymbol{z}$ to be internalized. Table \ref{tab:incontext_prompt_to_parameter_update} presents an overview of existing works.
\paragraph{Tackling Complex Tasks}
\citet{snell2022learning} show that context distillation is a general method to train LLMs. Specifically, they internalize three types of contexts, including abstract explanations, concrete examples, and step-by-step reasoning, using imitation learning (Eq. \ref{eq:ip2pu_IL}). Compared to in-context learning, it can use more training examples than the context window size allows via multiple rounds of context distillation. They also find that internalizing concrete training examples can outperform directly fine-tuning on the examples. \citet{deng2023implicit} propose a complicated framework called implicit chain-of-thought. Concretely, they train an emulator to predict the hidden states of the reasoning path when the reasoning path is absent and only the input is given, and a student model to read the output of the emulator for producing the output. They show that this approach can solve math reasoning tasks previously not solvable without explicit chain-of-thought, at a speed comparable to no chain-of-thought.

\paragraph{Model Customization}
\citet{askell2021general} find that it is sufficient to provide a long system prompt for transforming a pre-trained LLM into a moderately capable AI assistant. The prompt can be distilled into a new LLM using Eq. \ref{eq:ip2pu_KL}. This procedure is more effective than fine-tuning on the prompt. \citet{choi2022prompt} further show that the system prompt can be a long detailed text description of a persona or task instruction and study pseudo-input generation methods for building the training data (i.e., $\mathcal{D}$) in Eq. \ref{eq:ip2pu_IL}.
\paragraph{Knowledge Update}
This transformation can also be used to edit the knowledge inside an LLM. \citet{padmanabhan2024propagating} explore the injection of entity knowledge using context distillation. Their approach consists of two stages. At the first stage, the initial language model ($\pi_0$) is prompted to generate continuations from a textual definition of the target entity $\boldsymbol{z}$. At the second stage, the model parameters are updated so that the distribution of the new model ($\pi^*(\boldsymbol{y}|\boldsymbol{x}$) matches the distribution of the initial model conditioned on the definition ($\pi_0(\boldsymbol{y}|\boldsymbol{x}, \boldsymbol{z})$) on the transfer set ($\mathcal{D}$) (i.e., the objective defined in Eq. \ref{eq:ip2pu_KL}). They find this approach is more effective at making inferences based on injected facts than fine-tuning and other gradient-based knowledge-editing methods \citep{de-cao-etal-2021-editing,mitchell2022fast}.

\paragraph{Self-improving}
Another type of application of this transformation is to make an LLM to self-improve its reasoning ability without supervised data. For instance, \citet{huang2022large} first use the initial LLM to generate high-quality answers for unlabeled questions using advanced prompting strategies such as few-shot chain-of-thought prompting \citep{wei2022chain} and self-consistency \citep{wang2022self}, and fine-tune the LLM using those self-generated solutions as target outputs (Eq. \ref{eq:ip2pu_IL}). Their experiments show that this approach improves the results of the LLM across a wide range of reasoning tasks without any ground truth label.

\paragraph{Learning from Language Feedback}
In addition, \citet{scheurer2023training} propose imitation learning from language feedback (ILF), which applies the transformation to internalize language feedback. In ILF, the LLM is first instructed to generate multiple refined outputs given the input, the initial model output, and the external feedback. Then, the most feedback-incorporating refinement is selected and used to fine-tune the LLM (Eq.\ref{eq:ip2pu_IL}). This can be regarded as injecting informative language feedback into model parameters.

\subsection{Future Directions}
\paragraph{More Advanced Prompt Engineering}
One straightforward extension is that LLMs may benefit from internalizing more advanced prompt engineering methods \citep{shin2020autoprompt,zhou2022large,prasad2022grips,gonen2022demystifying,pryzant2023automatic,gao2023strategyllm}, including multi-agent interactions \citep{liang2023encouraging,du2023improving}, or a combination of different prompt engineering techniques (emulating the model being conditioned on multiple prompts simultaneously).

\paragraph{Learning from Concepts vs. Learning from Examples}
The great in-context learning capabilities of LLMs also make learning from concepts (e.g., learning to complete a task based on an abstract description or explanation of the task) an appealing alternative to the traditional paradigm of learning from examples (e.g., performing gradient descent on a set of input-output examples of a task). Learning from concepts not only saves the efforts of crafting or collecting training examples but also provides the flexibility for quick adaptations. For example, if there are some changes in the definition of a concept or desired behaviors, directly indicating such changes in the prompt is much more efficient and user-friendly \citep{akyurek-etal-2023-dune}. However, a rigorous performance comparison between learning from concepts and learning from examples is underexplored. The transformation from in-context prompt to parameter update can allow the model to learn from concepts in a more efficient and scalable fashion.

\paragraph{Lifelong In-context Learning}
A promising future for LLM-based AI assistants is their ability to memorize all interaction logs with users. By effectively processing and utilizing the information within the entire interaction history, the AI can make more informed decisions and offer more accurate and useful responses. Furthermore, this capability allows the AI to better understand user preferences and values, enabling it to personalize its interactions and deliver a more customized experience. However, due to the dual constraints of hardware resources and the model's ability to understand long texts~\citep{hsieh2024ruler}, it is clearly impractical to store the entire interaction history in within a prompt. To support lifelong in-context learning with a limited context input window, we believe that the transformation from in-context prompt to parameter update holds utmost significance. Periodic transformation acts as a vital link in ensuring the effective and efficient adaptability of the AI assistant over its lifetime. A related study \citep{wang2024greater} fine-tunes LLMs on previous generation history for long text generation tasks such as novel writing and discourse-level translation.
 
\paragraph{Direct Transformation with Hyper-networks}
Another interesting future direction is to invent better methods to complete the transformation. Existing approaches usually adopt an optimization process through context distillation~\citep{snell2022learning,padmanabhan2024propagating}, which remains a costly and tedious procedure, requiring careful preparations of training instances and tuning of optimization hyper-parameters. It would be more efficient and convenient if a hyper-network \citep{ha2017hypernetworks} could be constructed to directly convert in-context prompts into parameter updates, such as another neural network that takes the prompt as input and generates gradients or direct deltas to the model parameters. Previous work has explored hyper-networks for model editing with knowledge facts \citep{de-cao-etal-2021-editing,mitchell2022fast} and instruction tuning \citep{ivison2023hint}, but more comprehensive investigations on scalable transformation of arbitrary prompts are still lacking.

\section{Parameter Update $\Rightarrow$ In-Context Prompt}
\label{sec:pu2ip}
Following our discussion on the transformation from parameter update $\Rightarrow$ in-context prompt in \cref{sec:ip2pu}, another interesting direction is to find an in-context prompt $\boldsymbol{z}$ that can achieve the same effect as the parameter update $\pi_0 \rightarrow \pi^*$. Formally, we want to find $z$ given $\pi_0$ and $\pi^*$ such that
\begin{equation}
    \pi_0(\boldsymbol{y}|\boldsymbol{x}, \boldsymbol{z}) = \pi^*(\boldsymbol{y}|\boldsymbol{x}) \quad \forall \boldsymbol{x}, \boldsymbol{y}
    \label{eq:obj_param2icl}
\end{equation}

Similar to the transformation in \cref{sec:ip2pu}, this can be accomplished by knowledge distillation:
\begin{equation}
    \min _{\boldsymbol{z}} \mathbb{E}_{(\boldsymbol{x}, \boldsymbol{y})\sim \mathcal{D}} \mathbb{D}_{\text{KL}}\big[\pi_0(\boldsymbol{y} | \boldsymbol{x}, \boldsymbol{z}) || \pi^*(\boldsymbol{y} |\boldsymbol{x})  \big],
\end{equation}
or imitation learning:
\begin{equation}
    \max _{\boldsymbol{z}} \mathbb{E}_{\boldsymbol{x}\sim \mathcal{D}, \boldsymbol{y}\sim \pi^*(\boldsymbol{y}|\boldsymbol{x})} \pi_0(\boldsymbol{y} |\boldsymbol{x}, \boldsymbol{z}).
\end{equation}
However, the above optimization problems are much more difficult than their counterparts in Eq. \ref{eq:ip2pu_KL} and Eq. \ref{eq:ip2pu_IL} due to the non-differentiable nature of the concrete token sequence $\boldsymbol{z}$. One possible remedy is to optimize a sequence of continuous vectors instead of concrete tokens \citep{li2021prefixtuning,qin2021learninghowtoask,shin2020autoprompt,lester2021power,liu2023gptunderstand}. Another possible solution is to iteratively search and evaluate the best prompts \citep{prasad2022grips,zhou2022large,gonen2022demystifying,pryzant2023automatic}. It should be noted that if the parameter update $\pi_0 \rightarrow \pi^*$ is solely due to training on additional data and the training dataset $\mathcal{Z}$ is accessible, one could potentially encapsulate the entire training dataset within a prompt and utilize in-context learning $\pi_0(\boldsymbol{y} | \boldsymbol{x}, \mathcal{Z})$. This is becoming a practical approach as the context input windows of LLMs increase. However, this method does not guarantee the equivalent transformation in Eq. \ref{eq:obj_param2icl}.

\subsection{Applications}
To the best of our knowledge, this promising research line remains underexplored and has no direct applications in the literature. However, we find several related works worth mentioning. For example, \cite{Lin2024ReAlign} achieves effective alignment purely through in-context prompting with base LLMs. With just a few constant examples and a system prompt, prompting can match or even surpass the performance of LLMs aligned with supervised fine-tuning (SFT) and reinforcement learning from human feedback (RLHF). Nevertheless, their objective is not to mimic a particular parameter update $\pi_0 \rightarrow \pi^*$, notably deviating from our objective in Eq. \ref{eq:obj_param2icl}. Yet, their work suggests the great potential of human-readable prompts in adapting LLMs. Another interesting work is \citet{morris2023language}, which demonstrates that it is possible to recover the model input from the LLM's next-token distribution. Specifically, they train an inversion model of the encoder-decoder architecture to predict the prompt given the next-token probabilities. Their results confirm the feasibility of predicting prompts from model behaviors.

\subsection{Future Directions}
\label{sec:prompt2param_future}

Despite few works in the research line, we believe that converting parameter update $\pi_0\rightarrow\pi^*$ to in-context prompt $\boldsymbol{z}$ holds substantial potential in real scenarios.

\paragraph{Enhancing Inference Extensibility} One significant potential benefit is to enhance the extensibility of an LLM for various purposes (different tasks or domains). When serving an LLM $\pi_0$ for $K$ purposes, the fine-tuning-based methods need to store $K$ parameter updates for those purposes, leading to significant memory cost. Moreover, due to non-uniform parameters, the inability to concurrently decode examples of different purposes limits the use of model serving techniques (e.g., static or dynamic batching \citep{kwon2023efficient}), thereby compromising efficiency. In contrast, in-context prompts can naturally enable us to store a single copy of parameters in GPU memory and flexibly serve it for various purposes simultaneously by employing different in-context prompts.%This also enables mixed-task batches, where different examples in a batch of data correspond to different tasks by using different contexts in the input.

\paragraph{Manipulation of In-context Prompt} Additionally, in-context prompts can be more easily manipulated than parameter update, resulting extensive practical benefits in real-world scenarios. Therefore, in future works, we can explore how to leverage this convenient property in downstream tasks. For example, since the in-context prompt $\boldsymbol{z}$ is described using text, it can serve as a unified protocol when sharing parameter updates across deep learning frameworks and programming languages. For example, simplifying the operations when $\pi_0 \rightarrow \pi^*$ is trained using PyTorch \citep{paszke2019pytorch}, and served by FasterTransformer\footnote{\url{https://github.com/NVIDIA/FasterTransformer}}. 
Besides, many well-studied techniques on text, e.g., compression algorithms, can be directly employed to the in-context prompt. 

\paragraph{Understanding Parameter Update} Moreover, the exported in-context prompt in text format sheds light on understanding the parameter update from the perspective of humans. For example, the transformation can be used to derive a textual description of the differences between two models.

\section{In-Context Prompt $\Rightarrow$ Reward Model}
\label{sec:ip2rm}
In \cref{sec:ip2pu}, we demonstrate that the impact of an in-context prompt can be condensed into a parameter update, which conserves the context window and potentially facilitates a deep integration of multiple prompts. In this section, we further show that the influence of prompts can be leveraged to build a reward model. The resultant reward model can serve both evaluation and training purposes. We outline three approaches to building a reward model using in-context prompts as follows.
\begin{itemize}[wide=0\parindent,topsep=0em]
\item \textbf{Direct Prompting.}
The LLM can directly function as a reward model to score output by incorporating a grading prompt (e.g., ``Please rate the response on a scale of 1 to 10, with a higher score indicating better response quality'').
\begin{equation}
    r(\boldsymbol{x},\boldsymbol{y}) = \pi_0(s|\boldsymbol{x}, \boldsymbol{y}, \boldsymbol{z}) 
    \label{eq:ip_rm1}
\end{equation}
where $r$ denotes the resulting reward model, $\boldsymbol{z}$ represents the grading prompt, and $s$ is the verbalized score generated by the LLM $\pi_0$ for the input-output pair $(\boldsymbol{x}, \boldsymbol{y})$.

\item \textbf{Contrastive Prompting.}
The LLM can be employed to create pairwise preference data using two contrasting prompts: a positive prompt $\boldsymbol{z}^+$ that encourages the LLM to produce more desirable output, and a negative prompt $\boldsymbol{z}^-$ that encourages the LLM to generate less desirable output. It is also possible that either $\boldsymbol{z}^+$ or $\boldsymbol{z}^-$ is empty. Given an input $\boldsymbol{x}$, we provide the LLM with $\boldsymbol{z}^+$ and $\boldsymbol{z}^-$ to elicit corresponding outputs $\boldsymbol{y}^+ \sim \pi_0(\boldsymbol{y}|\boldsymbol{x}, \boldsymbol{z}^+)$ and $\boldsymbol{y}^-\sim \pi_0(\boldsymbol{y}|\boldsymbol{x}, \boldsymbol{z}^-)$ respectively. Consequently, the obtained preference data $(\boldsymbol{x}, \boldsymbol{y}^+, \boldsymbol{y}^-)$ can be used to train a reward model.

\item \textbf{Contrastive Scoring.}
\textit{Direct Prompting} and \textit{Contrastive Prompting} assume the base LLM can correctly understand the in-context prompt and reflect it in the generated content, which places high demands on the instruction-following and generation capabilities of the base LLM \citep{saunders2022self,gao2023scaling}. Conversely, the changes in output distributions under different prompts might be more sensitive, thereby providing more subtle and precise signals. Specifically, we can utilize the generation probability differences under contrastive prompts (i.e., $\pi_0(\boldsymbol{y}|\boldsymbol{x}, \boldsymbol{z}^+)$ vs. $\pi_0(\boldsymbol{y}|\boldsymbol{x}, \boldsymbol{z}^-)$) to obtain sequence-level score data (e.g., $\frac{\pi_0(\boldsymbol{y}|\boldsymbol{x}, \boldsymbol{z}^+)}{\pi_0(\boldsymbol{y}|\boldsymbol{x}, \boldsymbol{z}^-)}$) or token-level score data (e.g., $\frac{\pi_0(y_t|\boldsymbol{y}_{<t}, \boldsymbol{x}, \boldsymbol{z}^+)}{\pi_0(y_t|\boldsymbol{y}_{<t}, \boldsymbol{x}, \boldsymbol{z}^-)}$). The underlying intuition is generation probability of a desirable output or token under the positive prompt $\boldsymbol{z}^+$ should be greater than that under the negative prompt $\boldsymbol{z}^-$, and vice versa.
%The LLM can also be applied to attain point-wise score data using two contrasting prompts, which allows for more fine-grained quality assessment of an input-output pair. 
\end{itemize}

The core principle of \textit{Contrastive Prompting} and \textit{Contrastive Scoring} is that in-context prompts can alter model behavior, with varying prompts inducing distinct directional changes. The indirect use of in-context prompts through the lens of the reward model offers several benefits. First, the strength of the effect of in-context prompts can be amplified or controlled. Second, the impacts of different prompts can be combined and exploited.
\subsection{Applications}
\paragraph{Annotation-free Evaluation}
A comprehensive and unbiased assessment of LLMs often necessitates human evaluation \citep{ouyang2022training}. Nevertheless, acquiring human evaluations can be both expensive and time-consuming. Therefore, it is appealing to explore the use of strong LLMs, such as GPT-4, to perform evaluations following the protocol described in \textit{Direct Prompting}. This endeavor has proven to be effective, exhibiting a strong agreement with human evaluators \citep{zhou2024lima,li2024leveraging,alpaca_eval}. To further enhance the reliability of the LLM-generated scores, some studies \citep{li2023generative,zheng2023judging,chen2023alpagasus} incorporate a chain-of-thought prompting strategy \citep{wei2022chain}, which requests the LLM to provide reasoning for its assessment before determining a final quality score. For instance, the in-context prompt $\boldsymbol{z}$ could be ``Please first provide a comprehensive judgment to justify your evaluation. Based on your judgment, rate the response''. \cite{lin-chen-2023-llm} assess a response from multiple dimensions, including appropriateness and relevance, and generate a corresponding score for each dimension. Moreover, the assessment results from a strong LLM can also be used to train a local reward model \citep{bai2022constitutional,lee2023rlaif,li2023generative}, which plays a critical role in reinforcement learning from AI feedback \citep{lee2023rlaif}.

\paragraph{Annotation-free Alignment}
RLCD \citep{yang2023rlcd} introduces a framework for aligning LLMs without relying on human annotation. Specifically, it starts with an unaligned LLM and pairs of contrastive prompts, utilizing an automatic pairwise preference data generation pipeline based on the concept of \textit{Contrastive Prompting}. By feeding a positive prompt $\boldsymbol{z}^+$ that encourages directional change toward a desired attribute (e.g, ``to be more harmless'') and a negative prompt $\boldsymbol{z}^-$ that does the opposite (e.g., ``to be harmful'') into the base LLM, two outputs $\boldsymbol{y}^+$ and $\boldsymbol{y}^-$ are generated respectively. Then, $\boldsymbol{y}^+$ is automatically labeled as preferred over $\boldsymbol{y}^-$. RLCD follows the standard RLHF pipeline by training a reward model from the preference data and employing this reward model to run PPO for aligning the base LLM. Drawing upon the idea of \textit{Contrastive Scoring}, \cite{liu2024direct} further refine the RCLD framework \citep{yang2023rlcd} by incorporating a self-rewarding score by comparing the generation probabilities of the two outputs under the contrastive prompt pairs:
\begin{equation}
    R(\boldsymbol{x}, \boldsymbol{y}) = \log  \frac{\pi_0(\boldsymbol{y}|\boldsymbol{x},\boldsymbol{z}^+)}{\pi_0(\boldsymbol{y}|\boldsymbol{x},\boldsymbol{z}^-)}
    \label{eq:contrastive_scoring}
\end{equation}
and utilize a revised DPO algorithm, incorporating the relative score difference $R(\boldsymbol{x}, \boldsymbol{y}^+) - R(\boldsymbol{x}, \boldsymbol{y}^-)$ between the two responses $\boldsymbol{y}^+$ and $\boldsymbol{y}^-$, to effectively align the base LLM.

\paragraph{Learning from Language Feedback}
RLVF \citep{stephan2024rlvf} proposes a method for adapting LLMs to follow verbal feedback (e.g., ``Don’t use emojis when drafting emails to my boss'') post-deployment. Although such verbal feedback can be directly integrated into prompts, it presents at least two challenges. First, this approach requires incorporating the feedback into all subsequent inputs. Second, as the quantity of feedback accumulates, the prompt becomes increasingly lengthy, making the inference both more expensive and less accurate. In RLVF, GPT-4 initially generates a set of inputs where the given feedback should apply, followed by the base LLM being instructed to produce outputs for these inputs. For each input-output pair, the base LLM is then prompted with the verbal feedback to revise its original output $\boldsymbol{y}^-$ into an improved one $\boldsymbol{y}^+$. This can be considered a special variant of \textit{Contrastive Prompting}, where the negative prompt $\boldsymbol{z}^-$ is empty and the positive prompt $\boldsymbol{z}^+$ is a request asking for refinement (e.g., ``Please refine the response by incorporating the provided feedback: \{response\} \{feedback\}.''). The revised response $\boldsymbol{y}^+$ and the original one $\boldsymbol{y}^-$ are subsequently used as preference data to train the base LLM.

CUT \citep{xu2023reasons} presents a method for aligning LLMs with language feedback. Specifically, for each input and the corresponding output generated by the base LLM, language feedback (i.e., judgment) is provided to highlight the weaknesses of the output. The judgment can be drafted by humans, a strong LLM (e.g., GPT-4), or the base LLM itself. The objective of aligning LLMs with judgments is to enable LLMs to retain appropriate behaviors while addressing the weaknesses to prevent future misbehavior. Following the idea of \textit{Contrastive Scoring}, CUT uses the judgment to construct a negative prompt $\boldsymbol{z}^-$, asking the LLMs to generate an output that intentionally elicits the judgment. By contrasting the generation probabilities of the original output under the negative prompt and a positive prompt $\boldsymbol{z}^+$ at each time step, this approach can obtain token-level scores distinguishing appropriate tokens from inappropriate ones correlated with the judgment. Based on these scores, we can employ conventional likelihood training for appropriate content and unlikelihood training for inappropriate content. This work diverges from previous work in two perspectives. First, it targets at incorporating case-by-case feedback instead of global requirements or feedback. Second, it provides more fine-grained token-level scores rather than sequence-level scores.
\subsection{Future Directions}
\paragraph{Unifying Language Models and Reward Models}
Integrating a process-level reward model during inference can significantly enhance the reasoning capabilities of LLMs by providing timely feedback on the generated content \citep{khanov2024alignment,liu2024dont}. Nevertheless, this integration necessitates running multiple models concurrently, which increases the computational overhead. By leveraging the transformation from in-context prompt into reward model, it is feasible to combine the generative language model and the reward model into a unified system. The exploration of unifying models with differing purposes has been underway, with the integration of language models and representation models serving as a recent example \citep{muennighoff2024generative}.
\paragraph{Learning from Interactions}
Learning from language feedback provides richer supervision signals beyond pairwise preferences. However, this approach still relies on users explicitly providing judgments or comments on the model's outputs, which might not be convenient or even feasible for non-expert users. To address this limitation, one promising avenue is to learn directly from natural interactions with users; In common scenarios where LLMs engage in multi-turn conversations \citep{zhao2024wildchat, zheng2024lmsyschatm}, valuable feedback is often implicitly captured through subsequent interactions. These interactions may indicate the helpfulness or other relevant characteristics of the model responses. Moreover, LLMs can also benefit from engaging with varied environments, such as code execution scenarios \citep{jimenez2024swebench}, gaming contexts \citep{wang2023voyager}, and interactions with other agents \citep{meta2022human, liu2023agentbench}. These interactions can often be automated, allowing models to learn without direct human oversight, potentially leading to significant advancements in model capabilities and even superalignment.

\section{Reward Model $\Rightarrow$ In-Context Prompt}
\label{sec:rm2ip}
We have discussed the transformation from reward model to parameter update (\cref{sec:rm2pu}) and the mutual transferability between parameter update and in-context prompt (\cref{sec:ip2pu} and \cref{sec:pu2ip}). This suggests that the transformation from reward model to in-context prompt is also a feasible and worthwhile endeavor. In a similar spirit to the topic in \cref{sec:rm2pu}, the objective of this transformation can be defined as finding an optimal in-context prompt $\boldsymbol{z}$ that maximizes the expected reward value from $r(\boldsymbol{x}, \boldsymbol{y})$:
\begin{equation}
\max _{\boldsymbol{z}} \mathbb{E}_{\boldsymbol{x}\sim \mathcal{D}, \boldsymbol{y}\sim \pi_0(\cdot|\boldsymbol{x}, \boldsymbol{z})}[r(\boldsymbol{x}, \boldsymbol{y})],
\label{eq:rm2ip_special}
\end{equation}
where $\mathcal{D}$ denotes an empirical input distribution. The transformation from reward model to in-context prompt represents the long-standing goal of prompt engineering \citep{Sahoo2024ASS}, which involves crafting effective prompts to enhance task performance (rewards) without altering model parameters. Because manual prompt design is time-consuming and demands expert knowledge, great efforts have been made to automate the prompt engineering process.

\begin{itemize}[wide=0\parindent,topsep=0em]
\item \textbf{Gradient-based Methods.}
Several studies have harnessed gradient-based techniques for prompt optimization. 
\citet{shin2020autoprompt} refine prompts through iterative token substitutions guided by output likelihood maximization. \citet{shi2022toward} sample prompts from projected Langevin dynamics and incorporate a fluency constraint. These approaches depend on labeled training data and access to model parameters for gradient computation.

\item \textbf{Gradient-free Methods.}
Conversely, gradient-free methods aim to find the optimal prompt through (iterative) exploration and scoring. During the exploration phase, they create a range of prompt candidates using various techniques such as token- or phrase-level editing \citep{prasad2022grips,zhan2024unveiling}, developing task-specific prompt generators \citep{deng2022rlprompt,diao2023blackbox}, and employing LLMs to rephrase prompts \citep{zhou2022large}, where error feedback \citep{pryzant2023automatic,wang2023promptagent,chen2024prompt} and optimization trajectories \citep{yang2024large,guo2024connecting} can be leveraged to guide the rephrasing. Additionally, the guiding prompt, which directs the LLM in revising existing prompts, can itself be optimized using LLMs \citep{fernando2023promptbreeder}. Subsequently, the best prompts are selected based on downstream task performance or other proxy metrics, such as perplexity \citep{gonen2022demystifying}.

Another line of research aims to learn continuous prompts instead of discrete prompts through gradient-based optimization \citep{li2021prefixtuning,qin2021learninghowtoask,lester2021power,liu2023gptunderstand}. However, these methods rely heavily on extensive training data and yield prompts that lack interpretability.
\end{itemize}
\paragraph{Reward Model $\Rightarrow$ In-Context Prompt + Parameter Update}
The naive goal in Eq. \ref{eq:rm2ip_special} solely pursues the output with the highest reward value, which does not fully exploit the flexibility and versatility of in-context prompting. A more ambitious and naturally extended objective is to control the expected reward of the output through prompts. More formally, we are seeking for a prompting function $z$ that maps a given scalar value $s$ to a textual prompt $\boldsymbol{z} = z(s)$ such that the expected reward value $r(\boldsymbol{x},\boldsymbol{y})$ is equal to $s$ when $\boldsymbol{y}$ is sampled from $p(\cdot|\boldsymbol{x}, \boldsymbol{z} = z(s))$. It is important to note that this new objective encompasses the one expressed in Eq. \ref{eq:rm2ip_special}, as we can always condition the language model on the highest reward value during inference. More strictly speaking, the transformation is from reward model to in-context prompt + parameter update. To the best of our knowledge, most existing studies employ a manually-designed prompt function $z$ and optimize the parameters of $\pi$ through gradient-based optimization, possibly due to the optimization problems associated with the discrete space of possible prompts. We introduce two approaches for the transformation as follows.

\begin{itemize}[wide=0\parindent,topsep=0em]
\item \textbf{Reward-conditioned Supervised Fine-tuning.}
Similar to conventional supervised fine-tuning, this method maximizes the likelihood of exemplar outputs. In light of the additional condition $\boldsymbol{z} = z(s)$, we can extend existing input-output pairs $(\boldsymbol{x}, \boldsymbol{y})$ to input-output-reward triples $(\boldsymbol{x}, \boldsymbol{y}, s = r(\boldsymbol{x}, \boldsymbol{y}))$ using the reward model for annotation, and adopt the following optimization objective:

\begin{equation}
    \max_{\pi, z}  \mathbb{E}_{ (\boldsymbol{x}, \boldsymbol{y}) \sim \mathcal{D}} [\pi\Big(\boldsymbol{y}\Big|\boldsymbol{x}, z\big(r(\boldsymbol{x},\boldsymbol{y})\big)\Big)],
    \label{eq:rm2ip_condition}
\end{equation}
where $\mathcal{D}$ is a collection of input-output pairs that can either be collected offline or generated by the LLM itself given only the inputs.

\item \textbf{Meta-reward Model.}
This approach defines a meta-reward model $\hat{r}(\boldsymbol{x}, \boldsymbol{y}, s)$ using the difference between the input reward value $s$ and the outcome reward value $r(\boldsymbol{x}, \boldsymbol{y})$:
\begin{equation}
    \hat{r}(\boldsymbol{x}, \boldsymbol{y}, s) =  - | s - r(\boldsymbol{x}, \boldsymbol{y}) |.
    \label{eq:meta-reward}
\end{equation}
Consequently, we can optimize the LLM with respect to the meta reward model as Eq. \ref{eq:maximize_reward} in \cref{sec:rm2pu}.
\end{itemize}
\subsection{Applications}
\paragraph{Prompt Engineering}
%Existing approaches have been demonstrated to be effective across a variety of tasks, including standalone sentiment analysis \citep{shin2020autoprompt,shi2022toward,zhan2024unveiling,guo2024connecting,deng2022rlprompt}, natural language inference \citep{shin2020autoprompt,wang2023promptagent,zhan2024unveiling}, information extraction \citep{shin2020autoprompt,wang2023promptagent}, topic classification \citep{shi2022toward,guo2024connecting,deng2022rlprompt}, subjectivity classification \citep{wang2023promptagent,guo2024connecting,deng2022rlprompt}, linguistic acceptability \citep{zhan2024unveiling}, paraphrase identification \citep{zhan2024unveiling}, instruction induction \citep{zhou2022large,fernando2023promptbreeder}, question answering \citep{zhou2022large,wang2023promptagent,zhan2024unveiling}, arithmetic reasoning \citep{zhou2022large,fernando2023promptbreeder,yang2024large}, commonsense reasoning \citep{fernando2023promptbreeder}, toxic content detection \citep{pryzant2023automatic,fernando2023promptbreeder}, text summarization \citep{guo2024connecting}, text simplification \citep{guo2024connecting}, text style transfer \citep{deng2022rlprompt}, as well as comprehensive instruction-following tasks \citep{prasad2022grips,zhou2022large,wang2023promptagent,guo2024connecting,yang2024large} and agent-based multi-step tasks \citep{chen2024prompt}.
The best performance of current LLMs in diverse tasks relies on meticulously crafted prompts \citep{liu2023pre}. Studies reveal that minor prompt variations can significantly impact model performance \citep{gu2022robustness,mizrahi2023state,sun2023evaluating,sclar2023quantifying}, making prompt engineering a vital practice for LLM effectiveness. Existing approaches have been demonstrated to be effective across a variety of tasks, including specific NLP tasks, general-purpose instruction-following \citep{prasad2022grips,zhou2022large,wang2023promptagent,guo2024connecting,yang2024large}, and agent-based multi-step tasks \citep{chen2024prompt}.

Beyond unleashing the model's potential on downstream tasks, \citet{shin2020autoprompt} claim that prompt engineering acts as a valuable analytical instrument to evaluate the knowledge boundaries of the model. They also argue that, compared to fine-tuning, models can achieve higher average and worst-case performance in data-scarce situations, and are more efficient in terms of resource usage.
 
\paragraph{Unlearning on Negative Data}
LLMs may learn undesirable behaviors, such as offensive or toxic language, from large-scale corpora. It is therefore important to teach LLMs what not to do. To this end, \citet{lu2022quark} propose an unlearning approach using a reward model that quantifies the unwanted property. Their method scores LLM outputs with the reward model, prepends a reward token to the LLM's inputs based on the reward values, and optimizes the LLM to maximize the likelihood of the output given the augmented input (Eq. \ref{eq:rm2ip_condition}). At test time, the generations of the LLM are conditioned on the highest reward token, and evaluation results show that this framework is effective in unlearning toxicity, negative sentiment, and repetition. Similarly, \citet{zhang2023wisdom} introduce hindsight instruction relabeling, where the input $\boldsymbol{x}$ is augmented based on the correctness of the output $\boldsymbol{y}$, and show notable performance improvements on a range of reasoning tasks. \citet{korbak2023pretraining} suggest that the unlearning can also be conducted during the pre-training stage. Concretely, a reward model (e.g. a toxic text classifier) is used to classify pre-training text into two categories: good and bad. Each category corresponds to a special control token that is prepended to the text. In this way, the LLM can learn from undesirable content while being guided not to imitate it at inference time. \citet{liu2024chain} converts preference data into verbal comparisons. Given two outputs $\boldsymbol{y}^+$ and $\boldsymbol{y}^-$ for an input $\boldsymbol{x}$, where $\boldsymbol{y}^+$ is more preferred than $\boldsymbol{y}^-$, an equivalent expression can be constructed in natural language, such as ``Question: \{$\boldsymbol{x}$\} Good Answer: \{$\boldsymbol{y}^+$\} Bad Answer: \{$\boldsymbol{y}^-$\}''. Consequently, we can fine-tune language models on these verbal comparisons using the standard next-token prediction objective. At inference time, when prompted with a positive indicator such as ``Good Answer'', the model is expected to generate a desirable output.

\paragraph{Multi-dimensional Alignment}
To serve as powerful AI assistants, LLMs need to be aligned with a broad spectrum of human preferences and values. However, human preferences are inherently heterogeneous and multi-dimensional. The multifaceted nature of human preferences inadvertently introduces conflicts such as the dichotomy between harmlessness and helpfulness. Moreover, users may prioritize different dimensions in different application scenarios and training one model for each use case necessitates substantial computational resources. Therefore, it is appealing to control the LLM's preference settings through prompting. Following the above discussion, it is natural to explicitly specify the reward values for different reward models (e.g., helpfulness, harmlessness, and honesty) in a single prompt, thereby guiding the model to generate outputs that meet those expectations. \citet{guo2024controllable} and \citet{yang2024reward} combine the prompts from different reward models. For example, the resultant prompt $\boldsymbol{z}$ can be a sequence of special tokens such as ``<Helpfulness: 5> and <Harmlessness: 1>)''. For training, the idea of reward-conditioned supervised fine-tuning can be readily applied. In addition, \citet{guo2024controllable} introduce a variant of DPO \citep{rafailov2023direct} using the meta-reward model described in Eq. \ref{eq:meta-reward}. At inference time, different users may adjust the prompt to adapt to their personal desire on multiple alignment dimensions. 
 
\subsection{Future Directions}
\paragraph{Universal Prompt Engineering}
The current body of research on prompt engineering often segments prompts into two components: the \textit{task-level instruction} and the \textit{case-level input}. The task-level instruction serves as a broad outline and explanation of the task, while each case-level input provides specific details. 
Most studies have primarily concentrated on crafting the optimal task-level instructions, measuring model performance by averaging the results across various test cases. However, there are several shortcomings in this approach:
(\textit{i}) It fails to consider that the optimal task-level instruction may differ significantly from one case to another.
(\textit{ii}) It overlooks the substantial impact that rephrasing the case-level input might have on enhancing model performance.
(\textit{iii}) Real-world user queries seldom separate task-level instructions from case-level inputs distinctly and are typically not accompanied by a set of testing cases, rendering the traditional method of refining task-level instructions based on test-time model performance impractical.
In light of these insights, the development of a case-level prompt engineering strategy is essential, which should aim to create a universal prompt engineering function capable of adapting to the nuances of each case without evaluating on a test set. Such an approach is likely to be more effective in maximizing model performance by being inherently aware of the case-specific characteristics.

\paragraph{Control with Fine-grained Reward}
Many existing methods use discretized reward values, such as binary rewards for positive and negative data \citep{liu2024chain} or multi-level rewards from 1 to 5 \citep{guo2024controllable}. However, this discretization approach often overlooks detailed, fine-grained preference information, thereby limiting the model's ability to recognize subtle differences between outputs \citep{shen2023trickle}. As the main target of this transformation aims to enhance control over language model outputs, incorporating real-valued rewards or even language feedback from humans into this transformation could be more effective.

\section{Conclusions and Limitations}
In conclusion, our work provided a novel and comprehensive understanding of the interchangeability among parameter update, reward model, and in-context prompt in adapting pre-trained large language models (LLMs). By establishing a triangular framework with six transformation directions, we offer a unified view that connects diverse existing studies and suggests potential future research directions. This framework serves as a useful guide for researchers and practitioners in the field of LLMs, empowering them to make more informed decisions in their research and applications. Moreover, our work uncovers some new paths for exploring and harnessing the capabilities of LLMs.

We highlight several observations that may temper the popularity of the triangular framework. First, the mutual transformations between two elements are often imbalanced. While much of the research concentrates on one particular direction, there is scant attention paid to the reverse direction. Second, there is a deficiency in general and effective transformation methods for some transformation directions, which constrains the burgeon of many beneficial applications.
\bibliography{anthology, custom}
\bibliographystyle{acl_natbib}

\end{document}